\documentclass[10pt,twocolumn,letterpaper]{article}

\usepackage{cvpr}              % To produce the CAMERA-READY version
\definecolor{cvprblue}{rgb}{0.21,0.49,0.74}
\usepackage[pagebackref,breaklinks,colorlinks,allcolors=cvprblue]{hyperref}
\usepackage{algorithm}
\usepackage{algpseudocode}
\usepackage{amsmath}
\usepackage{makecell}
\usepackage{tabularx}
\usepackage{bbding}
\usepackage{amssymb}
\usepackage{xcolor}
\usepackage{colortbl}
\usepackage{graphicx}
\usepackage{adjustbox}
\usepackage{capt-of}
\usepackage{multirow}
\usepackage{multicol}
\usepackage{booktabs}
\usepackage{arydshln}
\usepackage{cuted}
\usepackage{tgadventor}
\usepackage{textcomp}

\usepackage{marvosym}
\definecolor{grey}{RGB}{240,240,240}
\definecolor{blue}{RGB}{217,232,242}
\definecolor{green}{RGB}{16,173,81}
\definecolor{red}{RGB}{175,37,25}
\def\paperID{30698} % *** Enter the Paper ID here
\def\confName{CVPR}
\def\confYear{2026}

\title{Pioneering Perceptual Video Fluency Assessment:\\ A Novel Task with Benchmark Dataset and Baseline}

\author{Qizhi Xie$^{1,2}$, Kun Yuan$^{2\textrm{ \Letter}}$, Yunpeng Qu$^{1,2}$, Ming Sun$^{2}$, Chao Zhou$^{2}$, Jihong Zhu$^{1\textrm{ \Letter}}$ \\
\textsuperscript{\rm 1} Tsinghua University,  \textsuperscript{\rm 2}Kuaishou Technology \\
{\tt \small xqz20@mail.tsinghua.edu.cn,}
{\tt \small yuankun03@kuaishou.com,}
{\tt \small jhzhu@tsinghua.edu.cn}
}
\begin{document}
\maketitle
\begin{abstract}
Accurately estimating humans' subjective feedback on \textbf{video fluency}, \eg, motion consistency and frame continuity, is crucial for various applications like streaming and gaming.
Yet, it has long been overlooked, as prior arts have focused on solving it in the video quality assessment (VQA) task, merely as a \textit{sub-dimension} of overall quality.
In this work, we conduct pilot experiments and reveal that current VQA predictions largely underrepresent fluency, thereby limiting their applicability.
To this end, we pioneer Video Fluency Assessment (VFA) as a standalone perceptual task focused on the temporal dimension.
To advance VFA research, \textbf{1)} we construct a fluency-oriented dataset, \textit{FluVid}, comprising 4,606 in-the-wild videos with balanced fluency distribution, featuring the first-ever scoring criteria and human study for VFA.
\textbf{2)} We develop a large-scale benchmark of 23 methods, the most comprehensive one thus far on FluVid, gathering insights for VFA-tailored model designs.
\textbf{3)} We propose a baseline model called \textit{FluNet}, which deploys temporal permuted self-attention (T-PSA) to enrich input fluency information and enhance long-range inter-frame interactions.
Our work not only achieves state-of-the-art performance but, more importantly, offers the community a roadmap to explore solutions for VFA.
Both FluVid and FluNet will be available at \url{https://github.com/KeiChiTse/VFA}.
\end{abstract}

\section{Introduction}

Video, as a representative multimedia medium, has experienced rapid growth in recent years, playing a pivotal role across various domains, such as social media~\cite{socialmedia1}, content creation~\cite{contentcreation}, and live e-commerce~\cite{ecom}.
According to Cisco's Visual Networking Index (VNI)~\cite{cisco}, global IP video traffic is expected to account for more than 82\% of all IP traffic by 2022.
Meanwhile, as video-capturing and viewing devices advance, users increasingly prefer \textit{videos that appear to be of ``high quality"}~\cite{vqa_qoe}.
The aforementioned trend poses a significant challenge for video providers in improving video quality to enhance users' Quality of Experience (QoE)~\cite{survey_qoe}.

\begin{figure}[t]
  \centering
  \includegraphics[width=\linewidth]{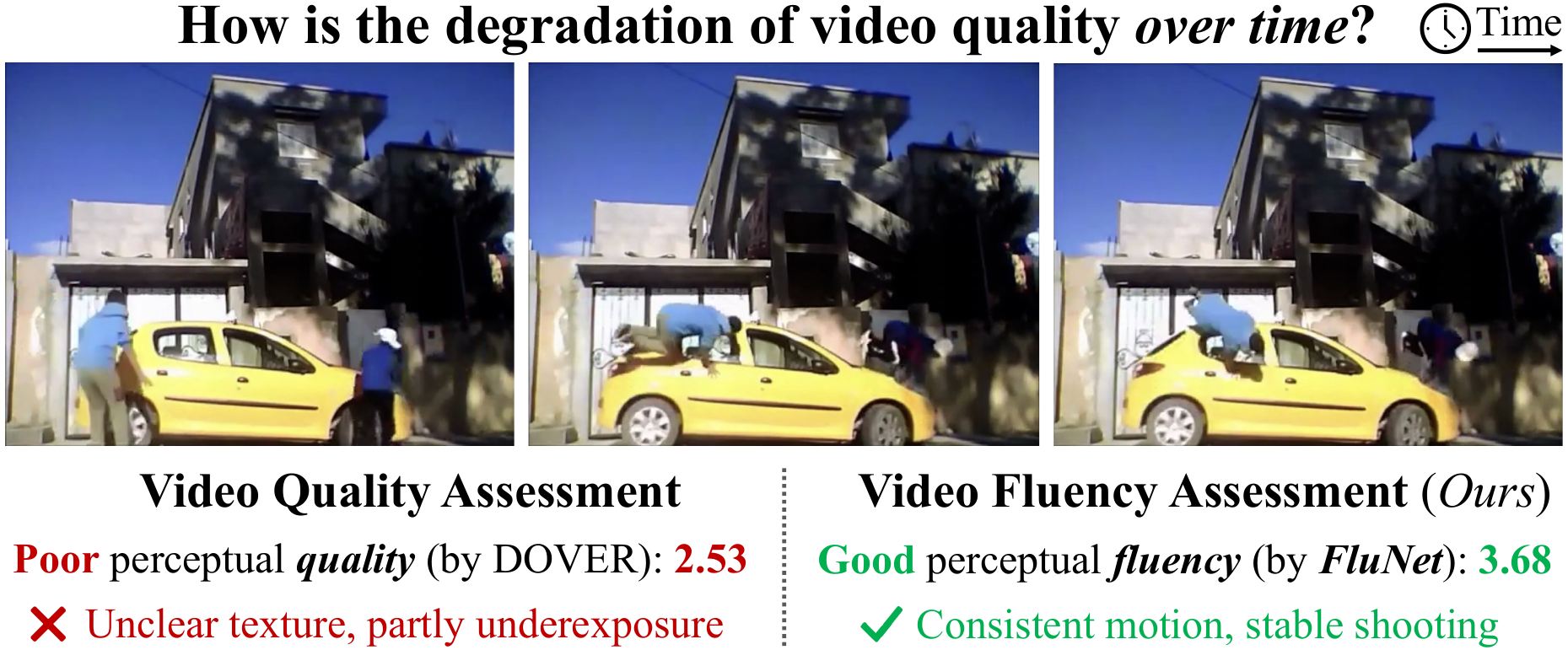}
  \captionsetup{font=small}
  \caption{{Motivation of video fluency assessment:} The prevailing VQA paradigm is highly sensitive to spatial distortions, whereas VFA concentrates exclusively on underrepresented temporal factors (\eg, camera motion, playback stuttering).}
  \vspace{-2mm}
  \label{fig:teaser}
\end{figure}

To tackle this challenge, the prevailing approach is to utilize off-the-shelf video quality assessment (VQA) models to quantify quality scores, thereby automatically emulating humans' subjective feedback on video quality~\cite{vsfa,simplevqa,pvq,vqt}.
As VQA research progresses~\cite{deepbvqa,maxwell,qbenchvideo}, numerous \textit{factors} are identified to affect human-perceived video quality, which can be categorized into \textit{frame-wise spatial} (\eg, noise, color, content composition \etc) and \textit{inter-frame temporal} distortions (\eg, screen shake, motion inconsistency \etc).
The coexistence of these factors renders VQA models' quantified scores as a \textit{holistic assessment} of both spatial and temporal quality, without explicitly disentangling them~\cite{reiqa,dover,cover}.
Inspired by prior studies~\cite{live,vmaf,livevqc} suggesting that \textit{human eyes are more sensitive to temporal than spatial distortion} when viewing a video, in this work, through a pioneering benchmark experiment (as in Tab.\ref{tab:benchmark_vqa}), we discover that the \textit{entangled prediction} of VQA models excels at identifying spatial \textit{rather than temporal quality} (as in Fig.\ref{fig:teaser}).
This prohibits the predicted scores from guiding temporal-related downstream tasks in video processing workflows, such as adaptive frame rate encoding~\cite{afe} and frame interpolation~\cite{vfi_cnn}.

To address these limitations of VQA experts, we refer to \textit{perceived temporal quality} as \textbf{fluency}, and propose a novel \textit{perceptual task}, named \textbf{Video Fluency Assessment (VFA)}, to emulate the human visual system's (HVS) perception of fluency accurately.
The task can be formulated as follows, where $V$ and $y$ denote the perceived video and its subjective fluency score, respectively:
\begin{equation}\label{equ:vfa}
    \begin{split}
    f:V\to y\\
    \end{split}
\end{equation}
The objective of VFA is to learn the \textit{mapping function} $f$ in Equ.~\ref{equ:vfa}.
More importantly, our VFA brings distinctive \textit{benefits}.
\textbf{First}, it can guide video quality enhancement more efficiently.
As in Fig.\ref{fig:teaser}, VFA prediction indicates subsequent efforts should focus on optimizing frame-wise clarity (\eg, using denoising operators) rather than fluency (\eg, conducting transcoding).
\textbf{Second}, VFA prediction can enhance text/image-to-video generation by serving as a fluency assessment metric or a reward for reinforcement learning.

In this paper, we dissect the proposed task and pinpoint three problems to advance VFA research:
\textbf{First}, there is a lack of a human subjective rating standard for video fluency.
As proved in VQA literature, a robust scoring standard—such as ITU-R BT.500~\cite {itu500}—serves as the cornerstone for defining subjective fluency levels and comparing the performance of different assessors~\cite{kv1k,livevqc}.
\textbf{Second}, the lack of large-scale human studies and datasets with ground-truth fluency scores $y$ presents a challenge, which not only hinders the benchmarking of various approaches but also constrains the full potential of data-driven methods empowered by deep learning~\cite{qpt,qptv2}.
\textbf{Last}, current VQA approaches lack explicit model designs to capture fluency, making them inadequate as baselines for the VFA task.

\textbf{To address the 1st problem}, we formulate a five-tier accurate category rating (ACR) \textit{criteria} to mapping human fluency perception (\eg, ``good") to numerical scores (\eg, 4.0), and vice versa.
The definition of five fluency levels is written by a group of 20 visual experts, strictly following the instructions in the ITU regulations~\cite{itu2022,itu91} for correctness.
\textbf{To address the 2nd problem}, we conduct the first-ever \textit{human study} for the VFA task using the proposed criteria, thereby constructing a \textit{dataset} termed \textbf{FluVid} for performance benchmarking.
FluVid comprises over 4,000 videos exhibiting diverse levels of fluency, each accompanied by a Mean Opinion Score (MOS) annotated by 20 experts. 
Moreover, to mitigate video selection bias and address the long-tailed fluency distribution for robust benchmarking, we meticulously devise a set of \textit{principles} to guide both video collection and annotation.
\textbf{To address the 3rd problem}, through a careful analysis of the performance of 23 models on FluVid, we ascertain that \textit{limited input frames} and \textit{insufficient inter-frame interactions} are the obstacles preventing them from perceiving fluency.
Therefore, we propose a simple yet powerful \textit{baseline model} named \textbf{FluNet}.
FluNet overcomes the obstacles by a computationally efficient \textit{attention mechanism} to compress and permute the channels of $\mathbf K$, $\mathbf V$ matrices.
Moreover, to address the reliance on expert-annotated fluency scores and leave FluVid for benchmarking purposes, we devise a self-supervised (SSL) \textit{training strategy}.
By synthesizing and learning to rank videos with various fluency levels, FluNet surpasses previous methods in VFA by a notable margin in a data-efficient manner.
Our main \textbf{contributions} are four-fold:
\begin{itemize}
    \item We pioneer an investigation into the task paradigm of VQA, revealing that current VQA models fall short in quantifying temporal quality. 
    To address this, we formulate temporal quality as fluency and present a novel perceptual \textbf{task}—Video Fluency Assessment (VFA).
    \item To the best of our knowledge, we are the first to establish the subjective fluency scoring standard and conduct a large-scale human study to construct the \textbf{dataset} \textit{FluVid}, covering over 4,000 videos and fluency-centred MOSs for performance benchmarking.
    \item We conduct a comprehensive \textbf{evaluation} on FluVid, covering 6 VQA methods and 17 open-sourced large multimodal models (LMMs) to measure their effectiveness in understanding video fluency.
    The results expose notable deficiencies in current methods and provide critical insights for future improvement.
    \item Driven by the analysis of benchmark results, we propose the \textbf{baseline model} \textit{FluNet} along with a ranking-based \textbf{training strategy}.
    Extensive experiments and ablation studies prove the effectiveness of our method.
\end{itemize}
\section{Related Work}
\textbf{Tasks.}
Accurately assessing perceptual fluency plays a critical role in multiple video tasks, including video quality assessment (VQA), video frame interpolation (VFI), and video stabilization (VS).
\textit{First}, VQA aims to evaluate the overall quality to mimic humans' feedback~\cite{ptmvqa,qptv2}. 
However, the predicted scores fail to disentangle spatial and temporal quality, and cannot serve as reliable indicators of fluency (as in Tab.\ref{tab:benchmark_vqa}).
In contrast, our proposed VFA yields accurate indicators for fluency.
\textit{Second}, VFI aims to generate intermediate frames between known frames to enhance motion coherence~\cite{emavfi,interpany}.
Assessing the change in fluency before and after interpolation is crucial.
Currently, VFI relies on pixel-based metrics, such as PSNR~\cite{itu81} and SSIM~\cite{ssim}, or VQA-based metrics, like LPIPS~\cite{lpips} and NIQE~\cite{niqe}.
Unlike these metrics, our proposed VFA method aligns well with HVS.
\textit{Last}, VS aims to eliminate unintended camera shakes to produce stable camera motion~\cite{faststab,survey_vs}.
Assessing the fluency of camera motion before and after stabilization is crucial.
Currently, VS estimates a stability score by computing the differential of the camera trajectory~\cite{opticalflow4vs}.
However, this score struggles to capture subtle changes in fluency~\cite{survey_ffs}, and our proposed VFA method can address this limitation for VS.

\noindent\textbf{Datasets.}
Numerous efforts in perceptual tasks (\eg, VQA) rely on time-consuming \textit{crowdsourcing} to construct large-scale datasets of human subjective scores (\eg, MOS), thereby enabling the training and benchmarking of models~\cite{kv1k,deepstab,snufilm}.
The data construction involves two steps: 1) video collection and 2) instructing annotators to score the videos by established criteria \cite{itu500}.
Recent VQA datasets exhibit features such as greater content diversity~\cite{maxwell}, larger scale~\cite{pvq}, and more annotation modalities~\cite{finevq}.
However, the above datasets focus on holistic video quality, lacking fluency-centered video collections and fine-grained annotations.
Therefore, they are not suitable for the VFA task and cannot support the generalization of VFA models.
To address this, we introduce \textit{FluVid} with the first fine-grained human study on fluency.
\begin{figure*}[t]
  \centering
  \captionsetup{font=small}
  \includegraphics[width=\linewidth]{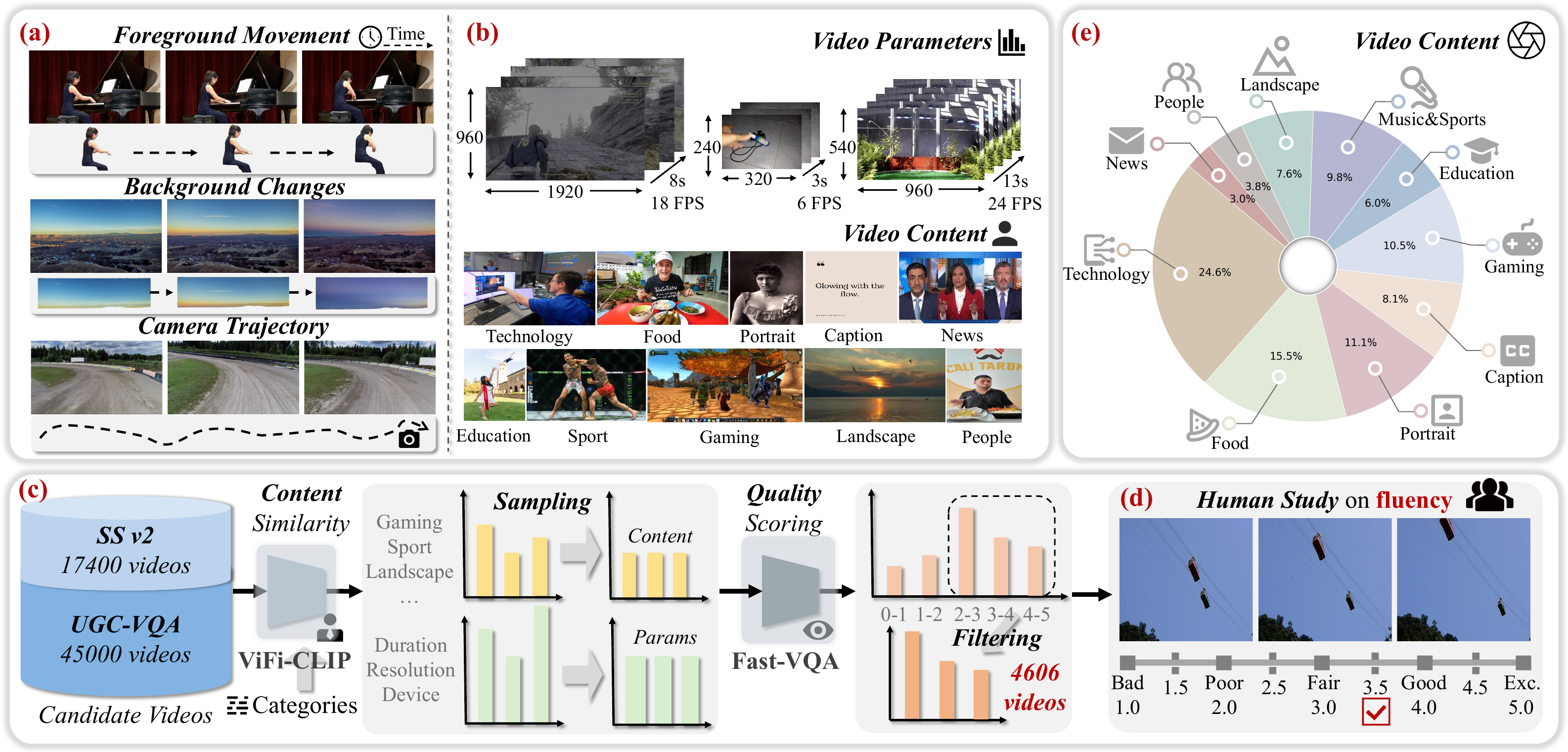}
  % \caption{\textbf{Construction pipeline of FluVid} includes: \textit{(a) }Principle 1: Incorporate three video components affecting fluency. \textit{(b) }Principle 2: Ensure video content and parameter diversity. \textit{(c) }A video curation and filtering process, guided by (a) and (b). \textit{(d) }A large-scale human study.}
  \caption{
  Construction pipeline of FluVid dataset. First, FluVid selects raw videos based on two guiding principles (a, b) to prevent selection bias and mitigate long-tail distribution effects. Second, leveraging the raw videos, we incorporate semantic information to remove duplicates, segment the content to ensure balanced distribution, and finally filter out segments exhibiting severe spatial distortions (c). Subsequently, we engaged 20 visual assessment experts to evaluate video fluency in a controlled laboratory setting (d), resulting in a dataset balanced across both categories and fluency.
  }
  \label{fig:data_pipeline}
\end{figure*}

\noindent\textbf{Methods.}
In the early stages, objective metrics such as stuttering rate and jank number are used to represent video fluency~\cite{fluency_objective_1,fluency_objective_2}. However, they cannot align with human subjective feedback~\cite{survey_qoe}. 
Concurrently, cognitive science research seeks to model the HVS perception of motion for fluency assessment~\cite{fluency_hvs_2,fluency_hvs_3}; however, these approaches overly simplify neural mechanisms and overlook individual variability~\cite{fluency_hvs_4}.
In recent years, data-driven approaches have driven remarkable progress in perceptual tasks (e.g., VQA), with a focus on generating more precise scores~\cite{vsfa,simplevqa,mdvqa,dover} or leveraging LMMs to produce linguistic descriptions~\cite{qinstruct,depictqa,qptv3}.
Nevertheless, these methods typically extract holistic quality features and lack tailored designs for perceiving subtle fluency variations, resulting in poor performance (as in Tab.\ref{tab:benchmark_vqa}).
To address this, we propose \textit{FluNet} to yield \textit{numerical scores} that accurately reflect perceptual fluency.

\section{The Benchmark Dataset: \textit{FluVid}}
Given the lack of quantitative evaluation, it is imperative to establish a benchmark for the VFA task.
% Therefore, this section presents the first-ever VFA benchmark, termed \textit{FluVid}.
This section elaborates on the construction pipeline and analysis of the first-ever VFA benchmark termed \textit{FluVid} as follows.
% are first established. 
% The video curation and subjective study are then illustrated following these principles. 
% The data analysis is lastly provided.
\subsection{Construction Principles}
To \textbf{avoid video selection bias} and \textbf{alleviate long-tailed fluency distribution} for reliable benchmarking, the FluVid is designed based on two guiding principles (Fig.\ref{fig:data_pipeline} (a)-(b)):
\begin{itemize}
    \item \textit{We specify three video components that dominate fluency and ensure the collected videos cover all of them} (Fig.\ref{fig:data_pipeline} (a)).
    Since interrelated factors (\eg, unstable shooting, dramatic light changes, \etc) affect video fluency, it is impractical to enumerate all factors and gather sufficient videos for each.
    Inspired by visual psychology and cognitive science research~\cite{camera,foreground,background}, we discover that all factors are governed by three video components: the \textbf{foreground}, the \textbf{background}, and the \textbf{camera}. 
    Therefore, we collect \textit{three categories} of videos, each with its fluency primarily determined by a specific video component. 
    This principle simplifies the collection while enriching diversity, allowing for a deeper understanding of VFA.
    \item \textit{We assure the diversity of video content and parameters} (Fig.\ref{fig:data_pipeline} (b)).
    Recent VQA advancements~\cite{vsfa,dover} suggest that fluency is related to \textbf{high-level semantics}. 
    Moreover, perceptual fluency is affected by \textbf{parameters} such as frame rate~\cite{youtubeugc,youtubeugc+}. 
    Therefore, FluVid collects videos with varied \textit{content categories} and \textit{parameters} to avoid selection bias.
\end{itemize}

\subsection{Video Curation}
\textbf{Source selection.}
As in Fig.\ref{fig:data_pipeline}(c), we select the action dataset SSv2~\cite{ssv2} as one of the sources, as it comprises videos in which the \textit{foreground movements}, such as the human body or objects, affect perceptual fluency (following \textbf{principle 1}).
Given its considerable size (over 220,000 videos), we randomly sample 100 videos each from 174 action classes, obtaining 17,400 videos.
In addition, we incorporate five UGC-VQA datasets as sources, including LSVQ~\cite{pvq}, KonViD-1k~\cite{kv1k}, LIVE-VQC~\cite{livevqc}, YouTube-UGC~\cite{youtubeugc}, and Maxwell~\cite{maxwell}. 
These datasets feature videos in which perceptual fluency stems from \textit{background changes} or \textit{camera trajectories} (following \textbf{principle 1}) and cover \textit{various parameters}, including resolution, duration, shooting devices, and shooting conditions (following \textbf{principle 2}). 
We sample from the UGC-VQA datasets and deduplicate to get 45,000 unique videos. 
To this end, a total of 624,000 videos are collected for the subsequent filtering.

\noindent\textbf{Filtering.}
As in Fig.\ref{fig:data_pipeline}(c), a filtering process is applied to ensure \textit{content and parameter diversity} (following \textbf{principle 2}).
Specifically, 10 distinct scene categories are selected from six source datasets. 
We leverage an off-the-shelf contrastive video captioner, ViFi-CLIP~\cite{vificlip}, to efficiently compute the similarity between each video and the 10 scene types, acquiring the content distribution of all videos. 
A uniform sampling is conducted on 624,000 videos to guarantee content and parameter diversity, yielding 7,648 videos.
In addition, numerous works indicate that \textit{severe spatial distortions} impair human judgment of video fluency~\cite{stgmsd,vmaf}. 
We exclude videos exhibiting such distortions from 7,648 videos to prevent a long-tailed fluency distribution. 
Specifically, we remove UGC-VQA videos with excessively low MOS and employ Fast-VQA~\cite{fastvqa} to predict the quality scores of videos from SSv2, discarding those with scores below 2.0, yielding \textbf{4,606} videos after filtering.

\subsection{Human Study} 
We present the first-ever human study as follows (Fig.\ref{fig:data_pipeline}(d), including the annotation criteria and guidelines to acquire MOS for 4,606 videos, which serves as the subjective gold standard for VFA.
%We organize the human study of VFA as follows (Fig.\ref{fig:data_pipeline}(d)):

\noindent\textbf{Criteria formulation.}
Inspired by the ITU standard followed in VQA human studies~\cite{mdvqa,maxwell,kvq}, we formulate a \textit{five-tier ACR} (Absolute Category Rating) criteria on perceptual fluency.
Similar to~\cite{itu500}, the criteria comprise five tiers—``Bad", ``Poor", ``Fair", ``Good", and ``Excellent"—to assess video fluency.
10 experts write and double-check the definitions and descriptions for correctness.

\noindent\textbf{Annotation setup.}
The human study is conducted by 20 professional researchers specializing in video processing in an in-lab environment displayed on MAC screens.
We download all the videos to prevent rebuffering and stalling during the annotation process.
To enable reliable scoring, a consistent stimulus method is utilized, allowing for repeated viewing of the same video.
To enable the fine-grained evaluation capability, the scoring tier, ranging from ``Bad" to ``Excellent," is divided into 1-5 intervals of 0.5.

\noindent\textbf{Annotation workflow.}
To ensure annotation quality, researchers first study the definitions and descriptions of the scoring criteria. 
They then evaluate 100 anchor videos labeled by experts to become familiar with the criteria and ensure consistency among them all.
Finally, they conduct the annotation process on 4,606 videos in the benchmark.

\begin{figure}[t]
  \centering
  \captionsetup{font=small}
  \includegraphics[width=0.9\linewidth]{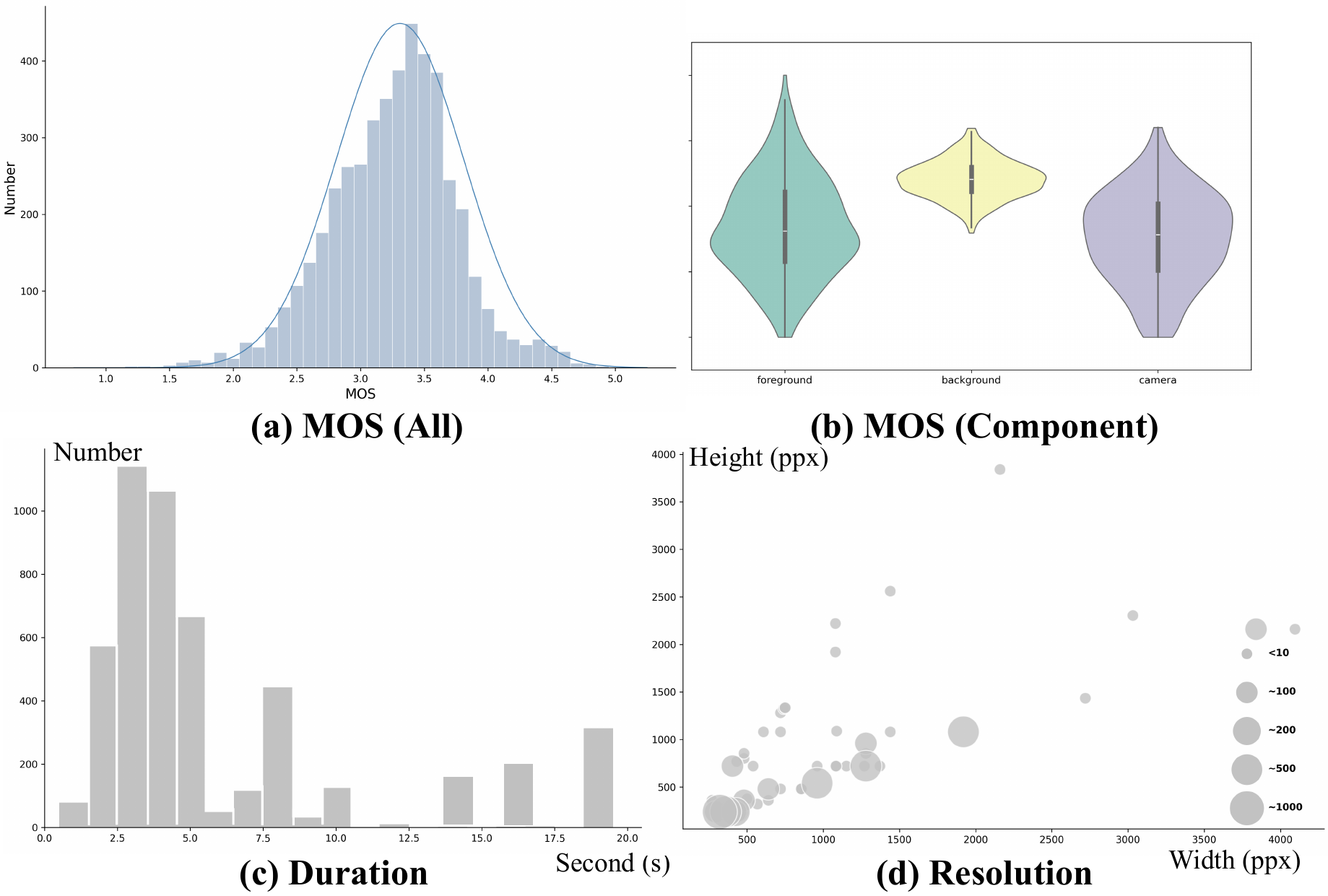}
  \caption{Statistical distribution of the FluVid dataset in terms of fluency, resolution, and duration.}
  \label{fig:data_stats}
\end{figure}
\begin{figure}[t]
  \centering
  \includegraphics[width=\linewidth]{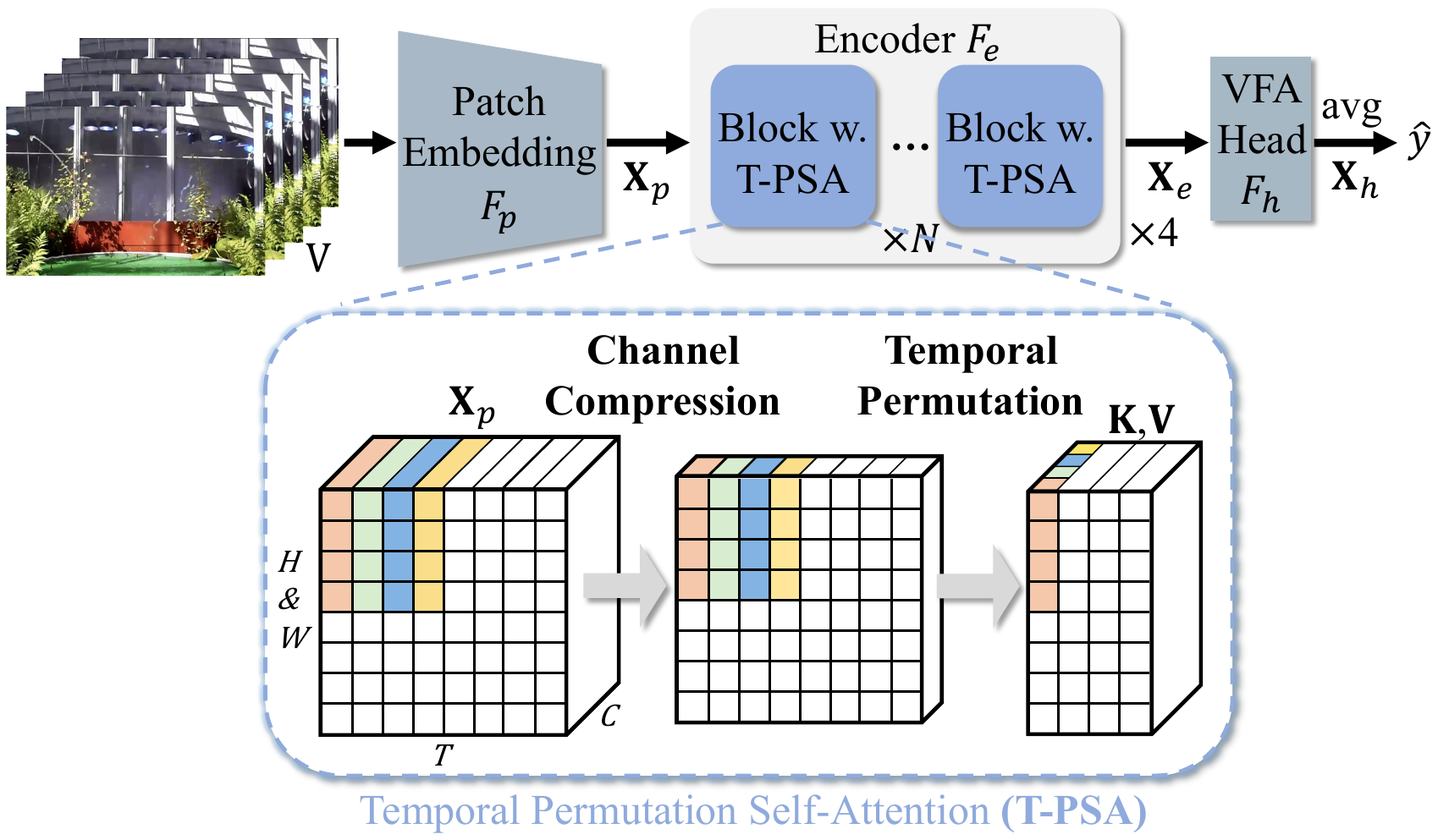}
  \captionsetup{font=small}
  % \caption{{Framework of FluNet: }It adopts a Swin-T-like architecture, T-PSA reduces the channel numbers and transfers the fluency information to the channel dimension to avoid information loss.}
  \caption{Framework of our proposed FluNet. By incorporating the T-PSA module for dimensional compression, FluNet achieves enhanced temporal perception capabilities.}
  \label{fig:model_pipeline}
\end{figure}
\subsection{Data Analysis}
We conduct a thorough analysis of the subjective fluency score for our FluVid in Fig.\ref{fig:data_pipeline}(c) and Fig.\ref{fig:data_stats}.
To investigate the impact of each video component, in addition to overall MOS, we also analyze the MOS distribution for three categories of videos (Fig.\ref{fig:data_stats} (b)), whose fluency is mainly influenced by foreground, background, and camera, respectively.
We observe that, compared to videos influenced by the background, those governed by foreground and camera movement tend to exhibit relatively \textit{lower} overall fluency scores.
This verifies motion from the filmed subject and the camera, as opposed to the background change, is more likely to result in stuttering.
Besides, the 10 scene categories in FluVid are evenly distributed, with the videos showcasing a rich diversity in both resolution and duration.
The above analysis is consistent with our construction principles, thereby proving the reliability of our construction pipeline and human study.
\section{The Baseline Model: \textit{FluNet}}
A potential reason that hinders VQA methods from perceiving fluency is the limited temporal information in the sparse frames (\eg, 16 frames).
Previous studies suggest that a direct solution is to increase the number of input frames or the frame rate~\cite{twostream,slowfast,videobert}.
% However, the computation burden prohibits practical training.
Inspired by~\cite{srformer}, we propose \textbf{temporal permuted self-attention (T-PSA)} to accommodate the \textit{increased input frames} and \textit{inter-frame interactions}, while balancing \textit{computation} simultaneously.
Given T-PSA, we construct a powerful yet simple \textbf{baseline model}, termed \textit{FluNet}.
Moreover, we propose an SSL \textbf{training strategy} to enhance FluNet's fluency perception capability and overcome the scarcity of fluency scores, described next.
% By learning to rank the synthesized videos with varying levels of fluency, the SSL training enhances FluNet's fluency perception capability and overcomes the scarcity of annotated fluency scores.

\subsection{Overall Architecture}
The architecture of FluNet is in Fig.\ref{fig:model_pipeline}, comprising three parts: a patch embedding layer $F_p$, an encoder $F_e$ with T-PSA transformer blocks, and a VFA head $F_h$.
Following previous works~\cite{swin,swin3d,fastvqa}, $F_p$ is a single convolution layer that first transforms the input video $V\in\mathbb R^{T\times H\times W\times 3}$ to feature map $\mathbf X_{p}\in\mathbb R^{\frac{T}{2}\times \frac{H}{4}\times \frac{W}{4}\times C}$, where $T$, $H$, $W$ denote the time, height, and width of the video, respectively, and $C$ denotes the channel dimension.
$X_{p}$ will then be sent to $F_e$ with a hierarchical structure. Similar to video swin transformer~\cite{swin3d}, $F_e$ includes four stages, each containing multiple T-PSA blocks.
After feature encoding, $\mathbf X_{e}\in\mathbb R^{\frac{T}{2}\times \frac{H}{32}\times \frac{W}{32}\times 8C}$ will be last sent to $F_h$ to regress the ground truth fluency scores $y\in\mathbb R$.
$F_h$ encompasses two point-wise convolution layers. 
The output feature map $\mathbf X_{h}\in\mathbb R^{\frac{T}{2}\times \frac{H}{32}\times \frac{W}{32}\times 1}$ is averaged along spatial and temporal dimensions to get the prediction $\hat{y}\in\mathbb R$.
\subsection{Temporal Permuted Self-Attention}
The core design of our FluNet is the T-PSA.
As in Fig.\ref{fig:model_pipeline}, given an input feature $X\in\mathbb R^{T\times H\times W\times C}$, we split it into $N$ non-overlapping windows $X\in\mathbb R^{NDS^2\times C}$, where $D$ and $S$ are the temporal and spatial side lengths of each window $W=(D,S,S)$.
Then, three linear layers $L_Q$, $L_K$, $L_V$ project $X$ to $\mathbf Q$, $\mathbf K$, $\mathbf V$ matrices as:
\begin{equation}\label{equ:qkv}
    \begin{split}
    \mathbf Q,\mathbf K,\mathbf V=L_Q(X),L_K(X),L_V(X)\\
    \end{split}
\end{equation}
Here, the weight of $L_Q$ retains the shape of $C\times C$, while $L_K$ and $L_V$ compress the weight matrices to $C\times\frac{C}{\gamma}$.
Thus, Equ.\ref{equ:qkv} yields $\mathbf{Q}\in\mathbb R^{NDS^2\times C}$, $\mathbf{K}\in\mathbb R^{NDS^2\times \frac{C}{\gamma}}$, and $\mathbf{V}\in\mathbb R^{NDS^2\times \frac{C}{\gamma}}$.
After this, we permute the temporal tokens in $\mathbf{K}$ and $\mathbf{V}$ to the channel dimension, obtaining $\mathbf{K}_p\in\mathbb R^{NS^2\frac{D}{\gamma}\times C}$, $\mathbf{V}_p\in\mathbb R^{NS^2\frac{D}{\gamma}\times C}$.
This permutation aligns the channel dimensions of $\mathbf{Q}$, $\mathbf{K}_p$, and $\mathbf{V}_p$ to compute self-attention.
Note that the window size of $\mathbf{K}_p$ and $\mathbf{V}_p$ are changed to $(\frac{D}{\gamma},S,S)$. 
Benefiting from the compression factor $\gamma$, T-PSA enables to \textit{increase the temporal side length} $D$ while keeping the computational cost, defined as:
\begin{equation}\label{equ:tpsa}
    \begin{split}
    \text{T-PSA}(\mathbf{Q},\mathbf{K}_p,\mathbf{V}_p)=\text{Softmax}(\frac{\mathbf{Q}\mathbf{K}_p^\top}{\sqrt{d_k}}+\mathbf{B})\mathbf{V}_p\\
    \end{split}
\end{equation}
Here, $\mathbf{B}$ is the relative positional bias~\cite{swin} and $\sqrt{d_k}$ is the scaling factor~\cite{trm}.

Our T-PSA poses unique advantages for the VFA task:
\textbf{First}, the permutation operation and the compressed channel $\frac{C}{\gamma}$ make it \textit{computationally efficient}. 
\textbf{Then}, it enables more input frames (\eg, $32$ to $128$) to provide \textit{richer fluency information} and a larger attention window size (\eg, $(8,7,7)$ to $(32,7,7)$) to achieve \textit{long-range inter-frame interactions}, leading to refined fluency modeling.
\textbf{Last}, unlike \cite{srformer}, T-PSA enlarges only the temporal window size $D$ while keeping the spatial size $S$ fixed, realizing a heightened focus on fluency than spatial details.
\begin{figure}
  \centering
  \includegraphics[width=\linewidth]{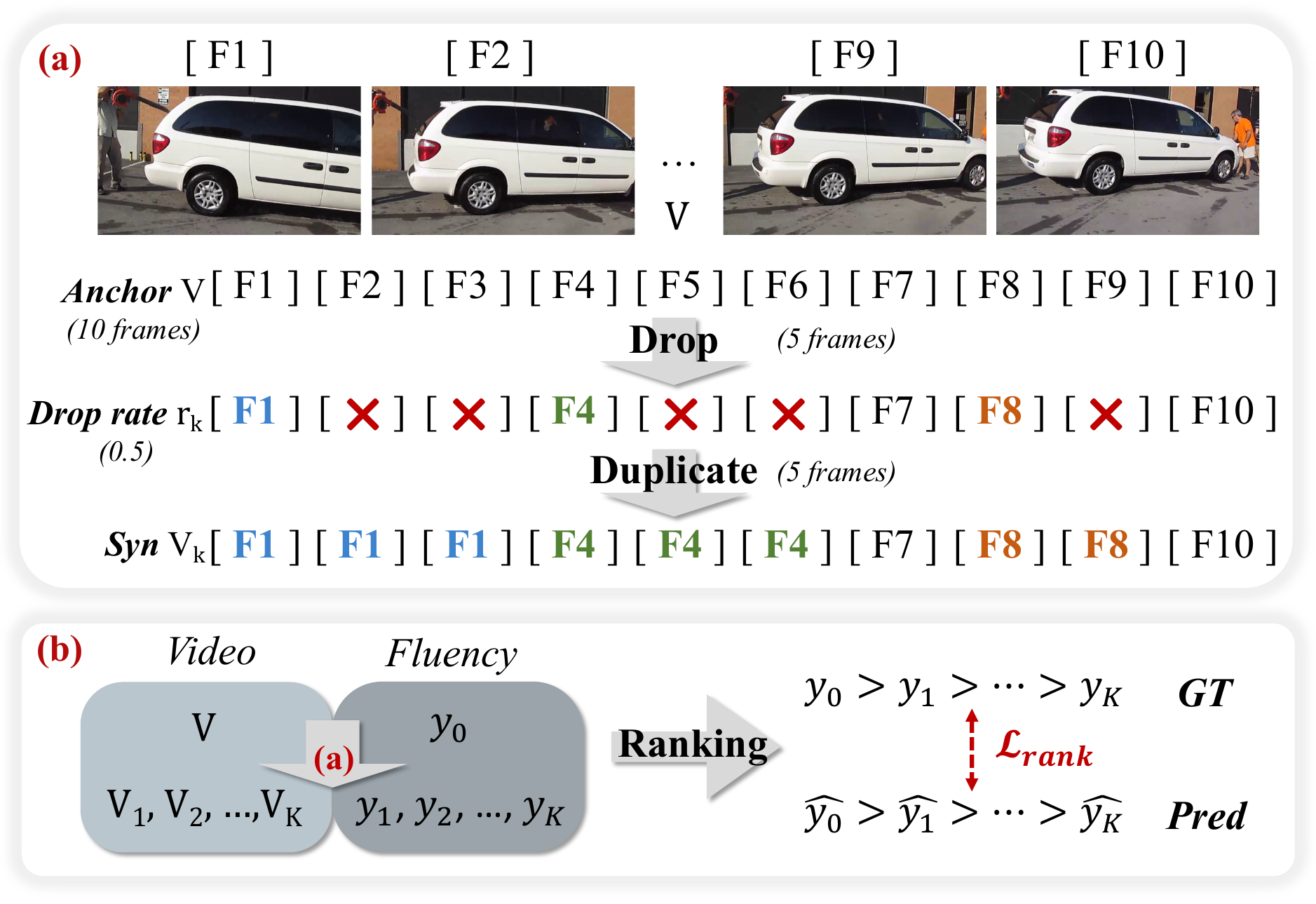}
  \captionsetup{font=small}
  % \caption{\textbf{Overview of training strategy: }\textit{(a)} Synthsize a stuttered video $V_k$ with anchor $V$ and drop rate $r_k$ by dropping and duplicating frames randomly. \textit{(b)} Conduct rank learning on $K+1$ videos.}
  \caption{Stuttered video simulation and Ranking Loss-Based optimization approach. By randomly dropping and padding video frames, we generate video sample pairs with varying levels of fluency, enabling the model to learn relative fluency judgments and thereby augment the number of training samples.}
  \label{fig:train}
\end{figure}

\subsection{Training Strategy}
The \textit{high cost} of human studies in VFA presents significant challenges to collecting scores~\cite{qptv2}.
Therefore, we propose an SSL strategy to leverage unlabeled videos and leave FluVid for benchmarking purposes.
The core idea is to \textit{enable FluNet to assess and rank videos of varying fluency levels}. Our training strategy is divided as follows (Fig.\ref{fig:train}):

\begin{algorithm}[t]
\caption{Synthesis of one of the $K$-ranked videos}
\label{alg:rank_syn}
\begin{algorithmic}[1]
\State \textbf{Input:} Anchor video $V$ of $T$ frames, drop rate $r_k$, number of intervals $M$
\State Compute total frames to drop: $T_{k}=T\times r_k$
\State Randomly divide $T_{k}$ into intervals: $T_{k_i}$ for $i = 1$ to $M$
\For{$i = 2$ to $M$}
    \State Compute remaining frames: 
    $T-\sum_{j=1}^{i-1} T_{k_j}$
    \State Randomly select a interval of length $T_{k_i}$
    % $[1, T-\sum_{j=1}^{i-1} T_{k_j}]$
    \State Generate \texttt{ffmpeg} command to drop and duplicate this interval
\EndFor
\State \textbf{Return:} a video of $T$ frames.
\end{algorithmic}
\end{algorithm}
\noindent\textbf{Synthesize ranked videos.}
To eliminate interference caused by spatial distortion, a total of 2,000 unlabeled, high-quality anchor videos are sampled from the HD-VILA dataset~\cite{hdvila}.
As in Fig.\ref{fig:train} (a), we synthesize $K$ videos for each anchor video.
The order of their fluency levels is controlled by \textit{dropping} and \textit{duplicating} different amounts of frames.
Also, to simulate real-world stuttering, frames are randomly dropped at various \textit{timestamps}.
%Alg.~\ref{alg:rank_syn} shows the pseudocode for synthesis.
We further shows pseudocode of the synthesis process in Alg.\ref{alg:rank_syn}.
To add more randomness in the synthesis, we introduce a hyperparameter $M$.
A total of $T_k$ frames will be randomly allocated to $M$ intervals, with the number of frames in each interval varying unpredictably.
We set $M$ to 5 in the paper.
In total, our synthesis step expands the training dataset to $\text{2,000}+\text{2,000}\times K$ ranked videos.
$K$ is set to 7, resulting in 16,000 videos for subsequent rank learning, described next.

% \begin{algorithm}
% \caption{Synthesis of one of the $K$ ranked videos}
% \label{alg:rank_syn}
% \begin{algorithmic}[1]
% \State \textbf{Input:} Anchor video $V$ of $T$ frames, drop rate $r_k$, number of intervals $M$
% \State Compute total frames to drop: $T_{k}=T\times r_k$
% \State Randomly divide $T_{k}$ into intervals: $T_{k_i}$ for $i = 1$ to $M$
% \For{$i = 2$ to $M$}
%     \State Compute remaining frames: 
%     $T-\sum_{j=1}^{i-1} T_{k_j}$
%     \State Randomly select a interval of length $T_{k_i}$
%     % $[1, T-\sum_{j=1}^{i-1} T_{k_j}]$
%     \State Generate \texttt{ffmpeg} command to drop and duplicate this interval
% \EndFor
% \State \textbf{Return:} a video of $T$ frames.
% \end{algorithmic}
% \end{algorithm}

\noindent \textbf{Train \textit{FluNet} for ranking.}
As illustrated in Fig.\ref{fig:train} (b), for each anchor video $V$ and its corresponding $K$ synthesised ranked videos $V_1$, ..., $V_K$, FluNet predicts their fluency scores as $\hat{y}_0$, $\hat{y}_1$, ..., $\hat{y}_K$.
The rank learning aims to discern the fluency order: $\hat{y}>\hat{y}_1>...>\hat{y}_K$.
Thus, for each anchor video, the optimization objective can be defined as:
\begin{equation}\label{equ:loss_r}
    \begin{split}
    \mathcal L_{\text{rank}}=\frac{1}{K}\sum_{i=0}^{K-1}\max(0, \hat{y}_{i+1}-\hat{y}_i+\beta)\\
    \end{split}
\end{equation}
$\beta$ is a margin that enforces a minimum separation between the ordered fluency levels and is set to 0.4.
After ranking extensive unlabeled videos, FluNet is fine-tuned on labeled videos to improve prediction accuracy, illustrated next.

\noindent \textbf{Fine-tune on VFA data.}
To boost the prediction accuracy of FluNet and calibrate its output scores, we sample 606 videos with fluency score $y$ from FluVid for fine-tuning, using the following $L_1$ loss: $\mathcal L_{\text{ft}}=\parallel\hat{y}_b-y_b\parallel_{1}$.

\begin{table*}[t]
\begin{minipage}{\textwidth}
\belowrulesep=0pt
\aboverulesep=0pt
\centering
\footnotesize
\captionsetup{font=small}
\caption{Performance benchmark of \textit{VQA methods} on the FluVid. The best and second-best results are \textbf{bolded} and \underline{underlined}.}
\vspace{-3mm}
\label{tab:benchmark_vqa}
\begin{tabular}{c|cccccc}
\toprule
\textbf{Metrics} & \makecell[c]{Fast-VQA\\(ECCV 2022)} & \makecell[c]{Faster-VQA\\(TPAMI 2023)} & \makecell[c]{DOVER\\(ICCV 2023)} & \makecell[c]{DOVER-M\\(ICCV 2023)} & \makecell[c]{SimpleVQA\\(MM 2022)} & \makecell[c]{PVQ\\(CVPR 2021)} \\
\midrule
SRCC$\uparrow$ & \textbf{0.640} & 0.627 & \underline{0.638} & 0.605 & 0.633 & 0.595 \\
PLCC$\uparrow$ & \textbf{0.633} & \underline{0.632} & 0.614 & 0.612 & 0.614 & 0.583 \\
\bottomrule
\end{tabular}
\end{minipage}

\vspace{3mm}
\begin{minipage}{\textwidth}
\belowrulesep=0pt
\aboverulesep=0pt
\centering
\footnotesize
\captionsetup{font=small}
\caption{Performance benchmark of \textit{video LMMs} on the FluVid. The best and second-best results are \textbf{bolded} and \underline{underlined}.}
\vspace{-3mm}
\label{tab:benchmark_lmm_vid}
\resizebox{1.0\textwidth}{!}{
\begin{tabular}{c|ccccccccc}
\toprule
\textbf{Metrics} & \makecell[c]{Video-LLaVA\\(ACL 2024)} & \makecell[c]{Chat-UniVi\\(CVPR 2024)} & \makecell[c]{Chat-UniVi-v1.5\\(CVPR 2024)} & \makecell[c]{LLaMA-VID\\(ECCV 2024)} & \makecell[c]{Video-ChatGPT\\(ACL 2024)} & \makecell[c]{PLLaVA\\(arXiv 2024)} & \makecell[c]{Qwen 2.5-VL\\(arXiv 2025)} & \makecell[c]{VQA$^2$-Scorer\\(arXiv 2025)} & \makecell[c]{Fine-VQ\\(CVPR 2025)} \\
\midrule
SRCC$\uparrow$ & 0.450 & 0.556 & 0.483 & 0.455 & 0.468 & 0.484 & 0.598 & \underline{0.602} & \textbf{0.622}\\
PLCC$\uparrow$ & 0.430 & 0.509 & 0.492 & 0.422 & 0.434 & 0.448 & 0.584 & \underline{0.603} & \textbf{0.609}\\
\bottomrule
\end{tabular}}
\end{minipage}

\vspace{3mm}
\begin{minipage}{\textwidth}
\belowrulesep=0pt
\aboverulesep=0pt
\centering
\footnotesize
\captionsetup{font=small}
\caption{Performance benchmark of \textit{image LMMs} on the FluVid. We evaluate Q-Align models trained on IQA, IAA, VQA, and all data, respectively. The best and second-best results are \textbf{bolded} and \underline{underlined}.}
\vspace{-3mm}
\label{tab:benchmark_lmm_img}
\begin{tabular}{c|ccccccccc}
\toprule
\textbf{Metrics} & \makecell[c]{Q-Align\\ (ICML 2024)} & \makecell[c]{Q-Align\\-IQA} & \makecell[c]{Q-Align\\-IAA} & \makecell[c]{Q-Align\\-VQA} & \makecell[c]{Q-Instruct\\ (CVPR 2024)} & \makecell[c]{ShareGPT4V\\(ECCV 2024)} & \makecell[c]{LLaVA-v1.5\\(CVPR 2024)} & \makecell[c]{LLaVA-OneVision\\(arXiv 2024)} \\
\midrule
SRCC$\uparrow$ & \textbf{0.568} & 0.503 & 0.493 & \underline{0.530} & 0.393 & 0.270 & 0.298 & 0.339 \\
PLCC$\uparrow$ & \underline{0.549} & 0.452 & 0.429 & \textbf{0.554} & 0.358 & 0.274 & 0.266 & 0.350 \\
\bottomrule
\end{tabular}
\end{minipage}
\vspace{3mm}

\begin{minipage}{\textwidth}
\belowrulesep=0pt
\aboverulesep=0pt
\centering
\footnotesize
\captionsetup{font=small}
\caption{Performance comparison of \textit{Fast-VQA} and our proposed \textit{FluNet} on the FluVid. We conduct rank learning and fine-tuning on Fast-VQA using its official checkpoint. The best and second-best results are \textbf{bolded} and \underline{underlined}.
}
\vspace{-3mm}
\label{tab:flunet}
\resizebox{0.98\textwidth}{!}{
\begin{tabular}{c|ccc|cccc|cc}
\toprule
\textbf{Methods} & Pre-training & Rank learning & Fine-tuning & Frames & Window size & GFLOPs & Params & SRCC$\uparrow$ & PLCC$\uparrow$ \\ \midrule
Fast-VQA & LSVQ & None & None & 32 & $(8,7,7)$ & 279 & 27.7M & 0.640 & 0.633 \\
(ECCV 2022) & LSVQ & None & None & 32 & $(16,7,7)$ & 148 & 27.7M & 0.626 & 0.607 \\
\textit{with larger window} & LSVQ & None & None & 64 & $(8,7,7)$ & 556 & 27.7M & 0.648 & 0.655 \\
\textit{or more frames} & LSVQ & None & None & 128 & $(8,7,7)$ & 1114 & 27.7M & 0.661 & 0.670 \\ \hline
Fast-VQA & LSVQ & \checkmark & None & 128 & $(8,7,7)$ & 1114 & 27.7M & 0.684 & 0.687 \\
\textit{with more frames} & LSVQ & None & \checkmark &  128 & $(8,7,7)$ & 1114 & 27.7M  & 0.669 & 0.650 \\
\textit{and further training} & LSVQ & \checkmark & \checkmark &  128 & $(8,7,7)$ & 1114 & 27.7M  & 0.725 & 0.716 \\ \midrule
\cellcolor{grey}\textbf{FluNet} (Ours) & None & \checkmark & \checkmark & 128 & $(32,7,7)$ & 308 & 26.5M & \underline{0.774} & \underline{0.770} \\
\cellcolor{grey}\textbf{FluNet++} (Ours) & LSVQ & \checkmark& \checkmark & 128 & $(32,7,7)$ & 308 & 26.5M & \textbf{0.816} & \textbf{0.821} \\ \bottomrule
\end{tabular}}
\end{minipage}

\end{table*}
\section{Experiments}
23 deep learning-based methods are \textit{first} benchmarked on FluVid to evaluate their performance.
One state-of-the-art method is \textit{then} retrained and compared with FluNet.
Extensive ablations are \textit{last} conducted to prove FluNet's efficacy.
\subsection{Benchmark on \textit{FluVid}}
\textbf{Model participants.}
For \textit{VQA methods}, we evaluate six models with two variants, including Fast-VQA~\cite{fastvqa}, Faster-VQA~\cite{fastervqa}, DOVER~\cite{dover}, PVQ~\cite{pvq}, and SimpleVQA~\cite{simplevqa}.
For LMMs, we evaluate nine \textit{video LMMs}, including Video-LLaVA~\cite{videollava}, Chat-UniVi-v1\&v1.5~\cite{chatuni}, LLaMA-VID~\cite{llama}, Video-ChatGPT~\cite{videochatgpt}, PLLaVA~\cite{pllava}, Qwen 2.5-VL~\cite{qwen2.5vl}, VQA$^2$-Scorer~\cite{qinstructvideo}, and FineVQ~\cite{finevq}, as well as eight \textit{image LMMs} with four variants, comprising ShareGPT4V~\cite{sharegpt4}, LLaVA-OneVision~\cite{llavaonevision}, LLaVA-v1.5~\cite{llava1.5}, Q-Instruct~\cite{qinstruct}, and Q-Align~\cite{qalign}.

\noindent\textbf{Evaluation settings.}
Our proposed VFA task adopts the SRCC and PLCC as the metrics.
Both metrics range in [-1, 1]; a larger SRCC indicates a better ranking between samples, and a larger PLCC shows a more accurate prediction.

For VQA methods, we use the final checkpoints to predict fluency scores.
For LMMs, we design an evaluation protocol to harness their power in predicting \textit{numerical scores}.
First, to avoid free-form output, the prompt is organized as follows, where \texttt{<img>} is the video token. 

\noindent\textit{\#User:} \texttt{<img>}\textit{How would you rate the fluency of this video?}
\textit{\#Assistant:} \textit{The fluency of the video is} \texttt{<level>}.

\noindent Then, a softmax-based strategy~\cite{qalign} is applied to transform \texttt{<level>} token to score.
Since \texttt{<level>} is the probability distribution on all possible tokens, including `bad', `poor', `fair', `good', and `excellent', denoted as $\mathcal X=\{\mathcal X_i|_{i=1}^5\}$.
Thus, we apply a close-set softmax to $\mathcal X$, and calculate the score $\hat{y}$ as a weighted average of 1-5:
\begin{equation}\label{equ:softmax}
    \begin{split}
    \hat{y}=\sum_{i=1}^{5}i\times\frac{\mathrm{e}^{\mathcal X_i}}{\sum_{j=1}^{5}\mathrm{e}^{\mathcal X_j}}\\
    \end{split}
\end{equation}

\noindent The protocol is implemented in the VLMEvalkit~\cite{vlmevalkit} codebase for all LMMs to ensure fairness.
Unless stated otherwise, the parameter count of all LMMs is 7B.

\noindent\textbf{Results.}
As in Tab.\ref{tab:benchmark_vqa}, the VQA methods achieve the most outstanding \textit{overall} performance across three categories.
It proves that accurately scoring the overall video quality contributes to enhancing the ability to assess fluency.
Nevertheless, as the focus of VQA and VFA differs, there remains considerable room for improvement.
Therefore, we embark on an exploration to enhance the fluency assessment capability of the SOTA VQA method, Fast-VQA (Tab.\ref{tab:flunet}).

In Tab.\ref{tab:benchmark_lmm_vid}, the overall performance of video LMMs ranks in the middle.
Specifically, seven general-purpose LMMs perform worse than the quality-aware VQA$^2$-Scorer and FineVQ, although the former excel in high-level vision benchmarks.
Among them, Qwen 2.5-VL demonstrates relatively better performance.
We believe its ability to process high-frame-rate videos with richer fluency information contributes to this.
In contrast, VQA$^2$-Scorer and FineVQ are fine-tuned on labeled quality data, yielding improved results in both SRCC and PLCC on FluVid.
Although FineVQ is fine-tuned on videos with fluency scores, we believe the limited fluency diversity (\eg, only user-generated content) and the vanilla model design lead to its degraded performance on FluVid. 
This further underscores the intrinsic connection between VQA and VFA tasks.

In Tab.\ref{tab:benchmark_lmm_img}, the overall performance of image LMMs trails behind.
As models conduct frame-by-frame inference and average the results to obtain the video fluency scores, we believe this renders them less capable of perceiving temporal dynamics to assess fluency. 
Still, similar to video LMMs, the general-purpose models exhibit relatively weak performance, whereas the quality-aware Q-Align performs better.

\subsection{Compare \textit{FluNet} with State-of-the-Art}

\textbf{Implementation details.}
During training and inference, FluVid takes a clip of 128 continuous frames, and each is resized to 224$\times$224.
For T-PSA, the window size is $(32,7,7)$, channel dimension $C$ is 96, and $\gamma$ is 2.
The four stages of the encoder $F_e$ have $(2,2,6,2)$ blocks, respectively.
$F_e$ is initialized by Swin-T~\cite{swin3d} pre-trained on Kinetics-400~\cite{k400}.
For rank learning and fine-tuning, experiments are conducted on eight NVIDIA V100 GPUs.
The batch size is 16.
AdamW is adopted with a weight decay of 0.01 and 0.05, and training lasts for 30 and 60 epochs, respectively.
The learning rates are 3e-4 and 1e-5, respectively, and decayed by cosine annealing without warmup.

\noindent\textbf{Results.}
Motivated by the findings in the benchmark results, we first push Fast-VQA's fluency assessment ability to compare it fairly with FluNet.
In Tab.\ref{tab:flunet}, the performance of Fast-VQA shows consistent improvement by increasing input frames.
However, enlarging the temporal length of the window hampers performance.
We think this outcome is due to the window mismatch with the pre-trained model.
Hence, we increase Fast-VQA's input to 128 frames and keep its window size unchanged for later training, including rank learning and fine-tuning.
Results in Tab.\ref{tab:flunet} prove the efficacy of further training: +6.4\% in SRCC and +4.6\% in PLCC when involving both stages.
Yet, the proposed FluNet baseline surpasses the improved Fast-VQA by a significant margin: \textbf{+4.9\%} in SRCC, \textbf{+5.4\%} in PLCC.
The computationally efficient design of FluNet empowers it to expand the window size for long-range inter-frame interactions, while keeping the GFLOPs within an affordable range.
Moreover, by pre-training FluNet on the LSVQ dataset beforehand to make it quality-aware, we propose FluNet++, which leverages the correlation between quality and fluency ratings and further advances its performance: \textbf{+4.2\%} in SRCC to 0.816, \textbf{+5.1\%} in PLCC to 0.821, showing its superior capability.

\begin{table}[t]
\begin{minipage}{\columnwidth}
\belowrulesep=0pt
\aboverulesep=0pt
\centering
\footnotesize
\captionsetup{font=small}
\captionof{table}{\textbf{Ablation study on \textit{FluNet} (I):} The proposed T-PSA, with all settings keeping the window size of $(32, 7, 7)$.}
\vspace{-3mm}
\label{tab:ablate_psa}
\scalebox{0.95}{
\begin{tabular}{cccc|ccc|cc}
\toprule
\multicolumn{4}{c|}{\textbf{Stages with T-PSA}} & Max & \multirow{2}{*}{GFLOPs} & \multirow{2}{*}{Params} & \multirow{2}{*}{SRCC$\uparrow$} & \multirow{2}{*}{PLCC$\uparrow$} \\
1 & 2 & 3 & 4 & frames &  &  &  &  \\ 
\midrule
  &   &   &   & 64  & 167 & 27.7M & 0.706 & 0.710 \\
\checkmark &   &   &   & 96  & 283 & 27.4M & 0.735 & 0.730 \\
\checkmark & \checkmark &   &   & 128  & 385 & 27.1M & 0.748 & 0.742 \\
\checkmark & \checkmark & \checkmark &  & 128 & 346 & 26.8M & 0.779 & 0.766 \\
\rowcolor{grey}\checkmark & \checkmark & \checkmark & \checkmark & 128 & 308 & 26.5M & 0.774 & 0.770 \\ 
\bottomrule
\end{tabular}}
\end{minipage}

\vspace{3mm}

\begin{minipage}{\columnwidth}
\belowrulesep=0pt
\aboverulesep=0pt
\centering
\footnotesize
\captionsetup{font=small}
\captionof{table}{\textbf{Ablation study on \textit{FluNet} (II):} The impact of window size in T-PSA, with all settings using 128 input frames.}
\vspace{-3mm}
\label{tab:ablate_ws}
\begin{tabular}{c|cc|cc}
\toprule
\textbf{Window size} & GFLOPs & Params & SRCC$\uparrow$ & PLCC$\uparrow$ \\ 
\midrule
$(8, 7, 7)$   & 1065 & 26.5M & 0.736 & 0.722 \\
$(16, 7, 7)$  & 622 & 26.5M & 0.758 & 0.749 \\
\rowcolor{grey}$(32, 7, 7)$ & 308 & 26.5M & 0.774 & 0.770 \\ 
\bottomrule
\end{tabular}
\end{minipage}

\end{table}

\begin{table}[t]
\begin{minipage}{\columnwidth}
\belowrulesep=0pt
\aboverulesep=0pt
\centering
\footnotesize
\captionsetup{font=small}
\caption{\textbf{Ablation study on \textit{FluNet} (III):} The proposed training strategy, with all settings \textit{not} pre-trained on LSVQ.}
\vspace{-3mm}
\label{tab:ablate_train}
\begin{tabular}{cc|ccc}
\toprule
\multicolumn{2}{c|}{\textbf{Training strategy}} & \multirow{2}{*}{SRCC$\uparrow$} & \multirow{2}{*}{PLCC$\uparrow$} & \multirow{2}{*}{KRCC$\uparrow$} \\
Rank learning & Fine-tuning &  &  \\ \midrule
 & & 0.694 & 0.703 & 0.528 \\
\checkmark &  & 0.722 & 0.718 & 0.574\\
 & \checkmark & 0.710 & 0.693 & 0.551\\
\multicolumn{2}{c|}{\textit{joint training}} & 0.753 & 0.748 & 0.593 \\
\rowcolor{grey}\checkmark & \checkmark & 0.774 & 0.770 & 0.621 \\
\bottomrule
\end{tabular}
\end{minipage}

\vspace{3mm}
\begin{minipage}{\columnwidth}
\belowrulesep=0pt
\aboverulesep=0pt
\centering
\footnotesize
\captionsetup{font=small}
\caption{\textbf{Ablation study on \textit{FluNet} (IV):} The impact of synthesised fluency levels. $K$ denotes the number of synthesized fluency levels of each anchor video in rank learning.}
\vspace{-3mm}
\begin{tabular}{c|c|cc}
\toprule
$K$ & \textbf{Drop rate} & SRCC$\uparrow$ & PLCC$\uparrow$ \\
\midrule
3 & $r_1$=0.1, $r_2$=0.5, $r_3$=0.9  & 0.726 & 0.732 \\
\midrule
\multirow{2}{*}{5} & $r_1$=0.1, $r_2$=0.3, $r_3$=0.5, & \multirow{2}{*}{0.748} & \multirow{2}{*}{0.751}\\
  & $r_4$=0.7, $r_5$=0.9  \\
\midrule
\cellcolor{grey} & $r_1$=0.1, $r_2$=0.2, $r_3$=0.3, $r_4$=0.5, & \multirow{2}{*}{0.774} & \multirow{2}{*}{0.770}  \\
\multirow{-2}{*}{\cellcolor{grey}7} & $r_5$=0.7, $r_6$=0.8, $r_7$=0.9& \\ \midrule
& $r_1$=0.1, $r_2$=0.2, $r_3$=0.3, & \multirow{3}{*}{0.774} & \multirow{3}{*}{0.770}  \\ 
& $r_4$=0.4, $r_5$=0.5, $r_6$=0.6,& \\
\multirow{-3}{*}{9}& $r_7$=0.7, $r_8$=0.8, $r_9$=0.9\\
\bottomrule
\end{tabular}
\label{tab:ablate_num}
\end{minipage}

\end{table}
\subsection{Ablation Studies on \textit{FluNet}}
To validate the effectiveness of FluNet and analyze the core design choices, ablation studies are conducted as follows:

\noindent\textbf{Effectiveness of T-PSA.}
In Tab.\ref{tab:ablate_psa}, we discuss the effectiveness of T-PSA.
The $1^{st}$ row represents the vanilla video Swin-T, which takes a maximum of 64 frames using a window size of $(32,7,7)$.
With the same window size, T-PSA introduces channel compression to reduce GFLOPs. 
As T-PSA is implemented in more blocks ( $2^{nd}$-$5^{th}$ rows), we find that the model processes additional frames, resulting in continuous performance improvements.
Moreover, with compressed channels of $\mathbf K$ and $\mathbf V$, T-PSA shows greater parameter efficiency compared to vanilla window self-attention.

\noindent \textbf{Impact of window size in T-PSA.}
The impact of window size is studied in Tab.\ref{tab:ablate_ws}.
The result in the $1^{st}$ row indicates that, although increased input frames encompass rich fluency information, a short temporal side length, as adopted by Fast-VQA, fails to model the overall fluency variations in the entire video. 
Instead, we believe it concentrates on the divergences between local frames.
Conversely, results in $2^{nd}$-$3^{rd}$ rows underscore the importance of simultaneously increasing both the window size and the number of input frames for improved understanding of fluency.

\noindent \textbf{Effectiveness of training strategy.}
The ablation of the training strategy is performed in Tab.\ref{tab:ablate_train}.
Comparing the $5^{th}$ row with the $1^{st}$-$3^{rd}$ rows, a clear conclusion can be drawn: both rank learning and fine-tuning boost fluency assessment.
Further, we compare the $4^{th}$ row with the joint training scheme ($3^{rd}$ row), which uses 606 labeled videos to synthesize ranked videos.
The training loss is a weighted sum of $\mathcal L_{\text{rank}}$ and $\mathcal L_{\text{ft}}$: $\mathcal L_{\text{ft}}+\alpha L_{\text{rank}}$, where $\alpha=0.3$.
Although joint training outperforms the results in the $1^{st}$ and $2^{nd}$ rows, the advantage of two-stage training lies in its ability to leverage massive unsupervised videos during rank learning. 
We believe the true potential of joint training will be revealed as the volume of fluency annotation continues to grow.

\noindent \textbf{Impact of synthesized fluency levels.}
In Tab.\ref{tab:ablate_num}, the effect of the number of synthesized fluency levels, $K$, used in rank learning is evaluated.
A larger $K$ allows for a more fine-grained fluency assessment, making the ranking task more challenging.
The best result can be acquired when $K$ is 7.

\section{Concluding Remarks and Future Work}
In this paper, we present a comprehensive study on the newly proposed VFA task by developing a baseline model, constructing a fluency-oriented benchmark dataset, and establishing the largest-scale benchmark to date. Extensive experiments and analyses underscore the significance of the VFA task. As a pioneering effort, future directions include scaling up fluency annotations, defining more fine-grained annotation criteria, and exploring more sophisticated model architectures. Overall, we hope this work will inspire the community to reevaluate current paradigms in perceptual assessment and contribute to the advancement of QoE.

{
    \small
    \bibliographystyle{ieeenat_fullname}
    \bibliography{main}
}

\clearpage
\setcounter{equation}{0}
\setcounter{figure}{0}
\setcounter{table}{0}
\setcounter{section}{0}
\renewcommand{\theequation}{\arabic{equation}}
\renewcommand{\thefigure}{\arabic{figure}}
\renewcommand{\thetable}{\arabic{table}}

\twocolumn[
\begin{center}
    {\Large \textbf{Pioneering Perceptual Video Fluency Assessment:\\ A Novel Task with Benchmark Dataset and Baseline \\ \vspace{0.5em} \textit{Supplementary Material}} \par}
    \vspace{1.5em}
    {\large
      \begin{tabular}[t]{c}
        Qizhi Xie$^{1,2}$, Kun Yuan$^{2\textrm{ \Letter}}$, Yunpeng Qu$^{1,2}$, Ming Sun$^2$, Chao Zhou$^2$, Jihong Zhu$^{1\textrm{ \Letter}}$ \\
        $^1$Tsinghua University, $^2$Kuaishou Technology \\
        \texttt{\small xqz20@mail.tsinghua.edu.cn, yuankun03@kuaishou.com, jhzhu@tsinghua.edu.cn}
      \end{tabular}\par
    }
    \vspace{2em}
\end{center}
]

\section{More Details of \textit{FluVid}}
\subsection{Scoring Criteria}
As described in the method section of the manuscript, we design a \textit{novel} five-point Absolute Category Rating (ACR) standard, meticulously adhering to the ITU regulations~\cite{itu81,itu2022,itu500,itu91}, designed to evaluate the \textbf{perceptual fluency} of 4,606 candidate videos.
It emphasizes the continuity of video across the temporal dimension, such as the stability of frame rates and the consistency of motion.
Here, we conclude the score and its corresponding visual descriptions and representative scenarios (double-checked and approved by 10 visual experts) in Tab.~\ref{tab:criteria}, which guide 20 annotators during the human study.
\subsection{Annotation Process}
\begin{figure*}[t]
  \centering
  \includegraphics[width=\linewidth]{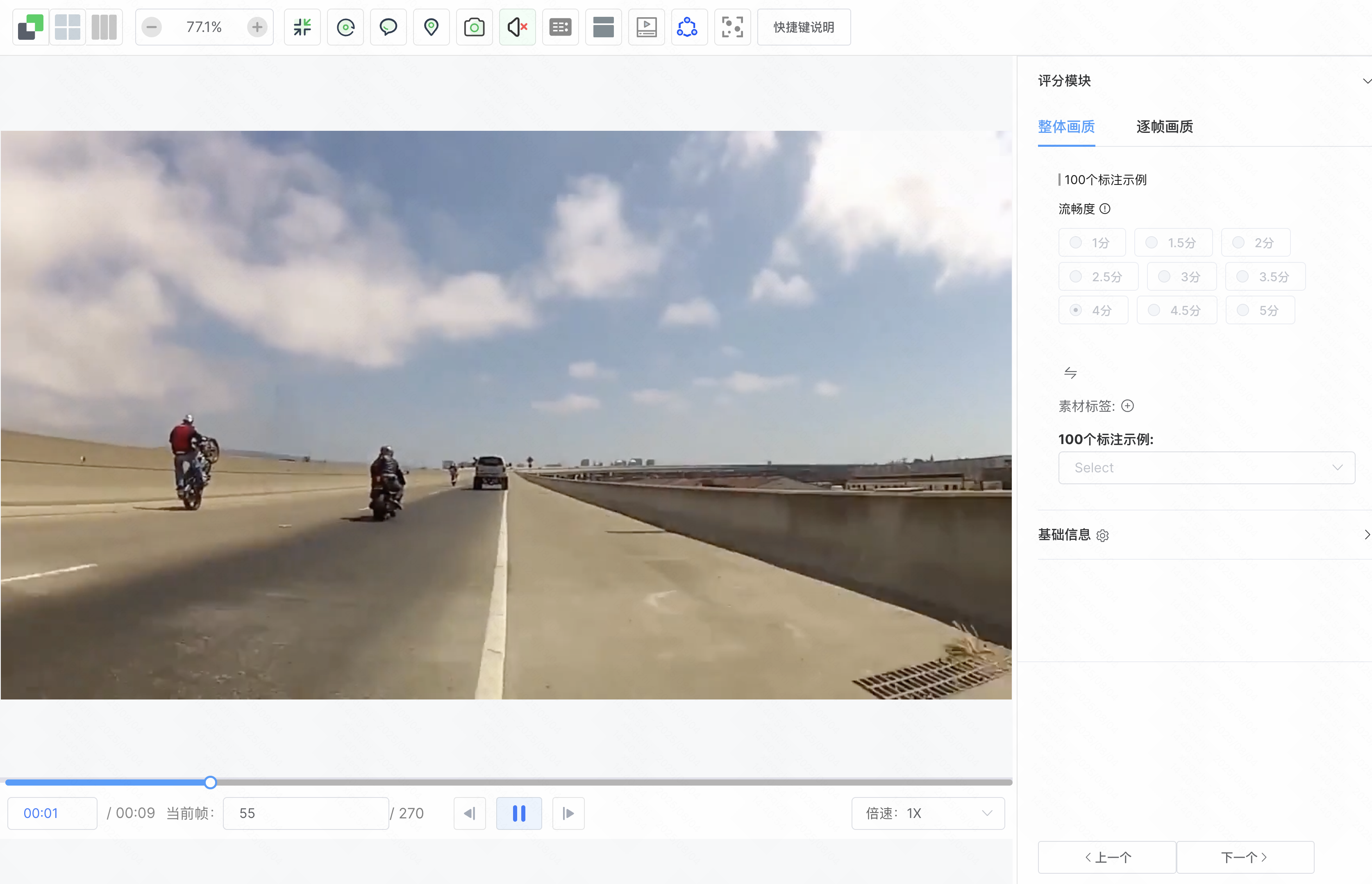}
  \captionsetup{font=small}
  \caption{Annotation GUI of human study.}
  \label{fig:gui}
\end{figure*}

In the human study section, we describe the annotation process of FluVid.
Here, we provide the annotation GUI as in Fig.\ref{fig:gui}.
On the left side, annotators need to watch the video playback thoroughly, focusing on the fluency of foreground objects, background, and camera movement.
Viewers are allowed to watch the playback repeatedly.
At the same time, annotators need to give their fluency rating on the right panel, following the criteria in Tab.\ref{tab:criteria}, as all of them are trained by scoring 100 anchor videos.
We collect the final scores of 20 annotators on each video and average them for the final results.
\subsection{Comparison to Existing Perceptual Datasets}
In the related work and method section, we highlight that FluVid is the first-ever perceptual benchmark dataset focused on fluency.
Here, we provide a comprehensive comparison among FluVid and other perceptual datasets (mostly in VQA task) in Tab.\ref{tab:dataset}.
% \begin{table*}[h]
% \centering
% \belowrulesep=0pt
% \aboverulesep=0pt
% \footnotesize
% \captionsetup{font=small}
% \caption{An overview of current public perceptual datasets.}
% \label{tab:dataset}
% \resizebox{1.0\textwidth}{!}{
% \begin{tabular}{ccccccc}
% \toprule
% \textbf{Dataset} & \textbf{Year} & \textbf{Duration/s} & \textbf{Ref. Num.} & \textbf{Scale} & \textbf{Scope} & \textbf{Subjective Evaluation Format} \\ \midrule
% CVD2014 & 2014 & 10-25 & - & 234 & In-capture VQA & In-lab \\
% Live-Qualcomm & 2016 & 15 & - & 208 & In-capture VQA & In-lab \\ \midrule
% KoNViD-1k & 2017 & 8 & - & 1,200 & In-the-wild VQA & Crowdsourced \\
% LIVE-VQC & 2018 & 10 & - & 585 & In-the-wild VQA & Crowdsourced \\
% YouTube-UGC & 2019 & 20 & - & 1,500 & In-the-wild VQA & Crowdsourced \\
% LSVQ & 2021 & 5-12 & - & 39,075 & In-the-wild VQA & Crowdsourced \\
% Maxwell & 2023 & 9 & - & 4,543 & In-the-wild VQA & Crowdsourced \\
% FineVQ & 2024 & 8 & - & 6104 & In-the-wild VQA & In-lab \\ \midrule
% UGC-VIDEO & 2019 & $>$10 & 50 & 550 & UGC w. compression & In-lab \\
% LIVE-WC & 2020 & 10 & 55 & 275 & UGC w. compression & In-lab \\
% YouTube-UGC+ & 2021 & 20 & 189 & 567 & UGC w. compression & In-lab \\ \midrule
% \rowcolor{grey}\textbf{\textit{FluVid} (Ours)} & 2025 & 2-20 & 100 & 4,606 & In-the-wild + UGC on VFA & In-lab\\ \bottomrule
% \end{tabular}}
% \end{table*}
We discover that most VQA datasets only provide holistic video quality scores (\eg, LIVE-VQC~\cite{live}, KoNViD-1k~\cite{kv1k} \etc).
While some datasets provide scores on one or more quality dimensions, such as DOVER~\cite{dover} and LSVQ~\cite{pvq}.
They do not disentangle fluency and pay insufficient attention to the temporal dimension.
Additionally, though Maxwell~\cite{maxwell} provides scores partly reflecting temporal quality (T-8), its scoring criteria are only a three-tier system, including a ternary choice of ``bad", ``neutral", and ``good".
This three-tier system is not fine-grained enough to align with the human visual system (HVS) and has not been proven by a large-scale human study to be as robust as our proposed five-tier rating.
FineVQ~\cite{finevq} constructs a VQA database FineVD, which includes temporal quality scores besides overall quality scores.
However, it focuses on addressing the VQA task, treating fluency merely as a sub-dimension of overall quality.
Conversely, we treat fluency as the sole focus of the newly proposed VFA.
Moreover, we conduct an in-depth analysis of fluency and dissect its factors into three video components, as in the method section.
Last but not least, the research focus of FineVD is on the language instructions for large multimodal models (LMMs).
\begin{table*}
    \begin{minipage}{\textwidth}
    \centering
    \belowrulesep=0pt
    \aboverulesep=0pt
    \footnotesize
    % \captionsetup{font=small}
    \caption{The annotation criteria for subjective labeling scores range from 1 to 5. Each scoring tier features its definition, visual description, and three representative scenarios. Please zoom in for a better view.}
    \label{tab:criteria}
    \resizebox{1.0\textwidth}{!}{
    \begin{tabular}{c|lll}
    \toprule
    \textbf{Score} & \multicolumn{1}{c}{\textbf{Definition}} & \multicolumn{1}{c}{\textbf{Visual Description}} & \multicolumn{1}{c}{\textbf{Representative Scenarios}} \\ \midrule
    1 Bad & \makecell[l]{The temporal sequence is severely\\ disrupted and incoherent.} & \makecell[l]{The temporal sequence is exceedingly choppy, marked by frequent and noticeable\\ stutters, frame drops, and drastic fluctuations in frame rate, rendering the actions\\ difficult to discern coherently and causing significant discomfort during viewing.\\} & \makecell[l]{1. Evident frame drops result in fragmented movements,\\ such as characters exhibiting a ``jumping" motion while walking;\\2. In fast-paced scenes, such as sporting events, severe\\ motion blur or screen tearing occurs, leading to discontinuous\\ object trajectories;\\3. In animated or gaming videos, characters display\\ unnatural pauses in their actions, with running movements\\ stuttering once every second.} \\
    2 Poor & \makecell[l]{The temporal sequence is noticeably\\ disjointed, hindering the viewing\\ experience.} & \makecell[l]{The temporal sequence exhibits clear stuttering and instability, with distortions\\ readily apparent; although one can barely discern the actions, the viewing\\ experience is frequently interrupted, diverting attention.} & \makecell[l]{1. Occasionally, frame duplication occurs, resulting in an \\alternating sequence of "stuttering and jumping" actions;\\2. In slow-motion scenes, such as when a character speaks,\\ there is a slight disjunction in head movements or gestures,\\ manifesting as 1 to 2 stutters per second;\\3. The rhythm of animated characters' movements is uneven,\\ with variations in stride length during walking, oscillating\\ between large and small steps.}\\
    3 Fair & \makecell[l]{The temporal sequence is moderately\\ smooth, with occasional imperfections.} & \makecell[l]{The temporal sequence is generally coherent, though it experiences occasional\\ fluctuations in fluidity; one must concentrate to perceive the imperfections, which\\ do not significantly hinder the overall comprehension of the content.} & \makecell[l]{1. In the long shot, there is a brief moment of stutter, manifesting\\ as a 0.5-second buffering trace;\\2. During rapid panning shots, the edges of background objects\\ exhibit a subtle, almost imperceptible\\ blurring discontinuity, as seen in the slight jerking of grass textures;\\3. In scenes with multiple characters, the movements of secondary\\ roles are slightly out of sync, such as the background actors lagging\\ just behind the rhythm of the protagonists.} \\
    4 Good & \makecell[l]{The temporal sequence flows gracefully,\\ approaching a natural cadence.} & \makecell[l]{The temporal sequence flows seamlessly and naturally, with virtually no stuttering\\ or fluctuations in frame rate; any temporal imperfections are scarcely perceptible\\ during viewing, and the continuity of movement approaches a professional filming\\ standard.} & \makecell[l]{1. In high-speed motion scenes, dynamic blur appears naturally, \\as exemplified by the seamless clarity of tire treads while racing;\\2. The intricate movements of complex actions, such as dance or\\ martial arts, flow together smoothly without any frame drops; \\3. During camera transitions or zooms, the shifts are fluid and\\ flicker-free, reminiscent of cinematic-grade editing.} \\
    5 Excellent & \makecell[l]{The sequence is exquisitely\\ fluid, without any imperfections.} & \makecell[l]{The temporal sequence is exquisitely coherent and smooth, with a perfectly stable\\ frame rate; the motion portrayal aligns seamlessly with real-world perception,\\ allowing for complete immersion in the content, free from any temporal \\distractions.} & \makecell[l]{1. There are no frame losses, repetitions, or misalignments; the\\ sequence of frames in the bitstream adheres strictly to the order of\\ capture.\\2. In ultra-high-speed motion scenes, such as the flight of a \\bullet or the collision of water droplets, each frame transitions\\ with precision, and the dynamic blur aligns with the laws of physics. \\3. In virtual reality (VR) videos, the visuals are rendered in real-time\\ without delay during head movements, effectively eliminating the\\ temporal sources of ``motion sickness."} \\ \bottomrule
    \end{tabular}}
    \end{minipage}
    
    \vspace{1em}
    
    \begin{minipage}{\textwidth}
    \centering
    \belowrulesep=0pt
    \aboverulesep=0pt
    \footnotesize
    % \captionsetup{font=small}
    \caption{An overview of current public perceptual datasets.}
    \label{tab:dataset}
    \resizebox{0.9\textwidth}{!}{
    \begin{tabular}{ccccccc}
    \toprule
    \textbf{Dataset} & \textbf{Year} & \textbf{Duration/s} & \textbf{Ref. Num.} & \textbf{Scale} & \textbf{Scope} & \textbf{Subjective Evaluation Format} \\ \midrule
    CVD2014 & 2014 & 10-25 & - & 234 & In-capture VQA & In-lab \\
    Live-Qualcomm & 2016 & 15 & - & 208 & In-capture VQA & In-lab \\ \midrule
    KoNViD-1k & 2017 & 8 & - & 1,200 & In-the-wild VQA & Crowdsourced \\
    LIVE-VQC & 2018 & 10 & - & 585 & In-the-wild VQA & Crowdsourced \\
    YouTube-UGC & 2019 & 20 & - & 1,500 & In-the-wild VQA & Crowdsourced \\
    LSVQ & 2021 & 5-12 & - & 39,075 & In-the-wild VQA & Crowdsourced \\
    Maxwell & 2023 & 9 & - & 4,543 & In-the-wild VQA & Crowdsourced \\
    FineVQ & 2024 & 8 & - & 6104 & In-the-wild VQA & In-lab \\ \midrule
    UGC-VIDEO & 2019 & $>$10 & 50 & 550 & UGC w. compression & In-lab \\
    LIVE-WC & 2020 & 10 & 55 & 275 & UGC w. compression & In-lab \\
    YouTube-UGC+ & 2021 & 20 & 189 & 567 & UGC w. compression & In-lab \\ \midrule
    \rowcolor{grey}\textbf{\textit{FluVid} (Ours)} & 2025 & 2-20 & 100 & 4,606 & In-the-wild + UGC on VFA & In-lab\\ \bottomrule
    \end{tabular}}
    \end{minipage}
\end{table*}

\section{More Details of \textit{FluNet}}
% \subsection{Synthesize Ranked Videos}
% In Fig.5 of the manuscript, we illustrate the training strategy.
% In part (a), we give an example for the synthesis of one of the $ K$-ranked videos.
% We further demonstrate the synthesis process in Alg.\ref{alg:rank_syn}.
% To add more randomness in the synthesis, we introduce a hyperparameter $M$.
% A total of $T_k$ frames will be randomly allocated to $M$ intervals, with the number of frames in each interval varying unpredictably.
% We set $M$ to five in the paper.
% \begin{algorithm}
% \caption{Synthesis of one of the $K$-ranked videos}
% \label{alg:rank_syn}
% \begin{algorithmic}[1]
% \State \textbf{Input:} Anchor video $V$ of $T$ frames, drop rate $r_k$, number of intervals $M$
% \State Compute total frames to drop: $T_{k}=T\times r_k$
% \State Randomly divide $T_{k}$ into intervals: $T_{k_i}$ for $i = 1$ to $M$
% \For{$i = 2$ to $M$}
%     \State Compute remaining frames: 
%     $T-\sum_{j=1}^{i-1} T_{k_j}$
%     \State Randomly select a interval of length $T_{k_i}$
%     % $[1, T-\sum_{j=1}^{i-1} T_{k_j}]$
%     \State Generate \texttt{ffmpeg} command to drop and duplicate this interval
% \EndFor
% \State \textbf{Return:} a video of $T$ frames.
% \end{algorithmic}
% \end{algorithm}
\subsection{Limitations of Synthesis and Mitigation}
In the method section, we emphasize that fluency is very complicated, sometimes dominated by foreground, background, or camera.
Still, we only drop and duplicate the whole frame when synthsizing stuttered videos, as in Fig.5 and Alg.1 of our paper.
A potential limitation for this ``drop-and-duplicate" strategy is that the synthesized stutter cannot cover the complex real-world scenarios.
To mitigate this, we further fine-tune FluNet on 606 videos with accurate fluency scores to enhance its generalization ability.
In future work, we will 1) introduce more sophisticated synthesis, such as applying the video segmentation techniques to change the fluency level of foreground, background, and camera, respectively. 2) Scale up FluVid to enable more videos for supervised training.
\subsection{Examples of Synthesized Videos}
To better illustrate our idea, we provide several exemplary synthesized videos with various drop rates in the supplementary material.
By modifying the drop rate $r_k$, we can control the fluency level in a fine-grained manner.
\section{More Details of Experiments}
\subsection{Rationale of Model Participant Selection}
In the experimental settings section, we briefly introduce 23 model participants.
Here, we explain the rationale behind our selection.
First, we select VQA methods based on their performance on common VQA benchmarks, including LSVQ, LIVE-VQC, KoNViD-1k and YouTube-UGC~\cite{youtubeugc}.
Thus, six state-of-the-art VQA methods are selected.
Second, large multimodal models (LMMs) begin to excel in multiple vision tasks, including high-level (\eg, classification, detection \etc) and low-level tasks (\eg, deblurring).
Therefore, it is imperative to benchmark them on our VFA task.
Among them, video LMMs are selected due to their superior ability to process videos.
Specifically, we choose seven video LMMs with best performance on general video benchmarks, including VideoMME~\cite{videomme}, MMBench-Video~\cite{mmbenchvideo}, MVBench~\cite{mvbench}, and MLVU~\cite{mlvu}.
We also choose two video LMMs specialized in VQA, including FineVQ and VQA$^2$-Scorer.
Recent literature suggests image LMMs exhibit primary quality perception ability. 
Hence, we select three image LMMs with top-tier performance on general image benchmarks, including MME~\cite{mme}, MMBench~\cite{mmbench}, MMMU~\cite{mmmu}, MTVQA~\cite{mtvqa}, and two image LMMs specialized in image quality assessment (IQA).
\subsection{Benchmark Settings}
In the experimental settings section, we state the detailed benchmark setting on FluVid, including evaluation prompts and the softmax strategy.
Here, we provide additional details for clarity.
In terms of the number of frames, we use the default setting of each model to activate the best performance for fairness.
For VQA models, the frame numbers for Fast-VQA~\cite{fastvqa}, Faster-VQA~\cite{fastervqa}, DOVER~\cite{dover}, SimpleVQA~\cite{simplevqa}, and PVQ~\cite{pvq} are 32, 16, 32, 32 and 16, respectively.
Note that, for SimpleVQA and PVQ, we pre-extract the motion features using the pre-trained SlowFast-ResNet-50~\cite{slowfast}.
For video LMMs, the frame numbers or frame rate for Video-LLaVA~\cite{videollava}, Chat-UniVi-v1\&v1.5~\cite{chatuni}, LLaMA-VID~\cite{llama}, Video-ChatGPT~\cite{videochatgpt}, PLLaVA~\cite{pllava}, Qwen 2.5-VL~\cite{qwen2.5vl}, VQA$^2$-Scorer~\cite{qinstructvideo}, and FineVQ~\cite{finevq} are eight, 64, 1FPS, 100, 16, 2FPS, 1FPS, and eight frames, respectively.
For all eight image LMMs, we uniformly sample 64 frames and average the per-frame scores for evaluation.

\end{document}